# Human heuristics for AI-generated language are flawed

Maurice Jakesch[a,b,*], Jeffrey T Hancock[c], Mor Naaman[a,b],

[a]*Cornell University,* [b]*Cornell Tech,* [c]*Stanford University*

[*]*Corresponding author email:* mpj32@cornell.edu

Human communication is increasingly intermixed with language generated by AI. Across chat, email, and social media, AI systems suggest words, complete sentences, or produce entire conversations. AI-generated language is often not identified as such but presented as language written by humans, raising concerns about novel forms of deception and manipulation. Here, we study how humans discern whether verbal self-presentations, one of the most personal and consequential forms of language, were generated by AI. In six experiments, participants (N = 4,600) were unable to detect self-presentations generated by state-of-the-art AI language models in professional, hospitality, and dating contexts. A computational analysis of language features shows that human judgments of AI-generated language are hindered by intuitive but flawed heuristics such as associating first-person pronouns, use of contractions, or family topics with human-written language. We experimentally demonstrate that these heuristics make human judgment of AI-generated language predictable and manipulable, allowing AI systems to produce text perceived as “more human than human.” We discuss solutions, such as AI accents, to reduce the deceptive potential of language generated by AI, limiting the subversion of human intuition.

**Author Contributions:** M.J. conducted the study, analyzed the results, and wrote the initial draft of the manuscript. M.J. and M.N. developed the study design and research framework. All authors contributed to the final manuscript.

**Competing Interest Statement:** The authors declare no conflict of interest.

**Funding information:** This material is based upon work supported by the National Science Foundation under Grant No. CHS 1901151/1901329 and the German National Academic Foundation.

## Introduction

Large generative language models (1, 2) produce semantic artifacts closely resembling language created by humans. Through applications like smart replies, writing autocompletion, grammatical assistance, and machine translation, AI-enabled systems infuse human communication with generated language at a massive scale. Large language models like OpenAI's GPT-3 and AI language applications like ChatGPT (1, 2) produce coherent writing pieces and generate entire conversations. AI-generated language enables novel interactions that reduce human effort but can facilitate novel forms of plagiarism, manipulation, and deception (1, 3–8) when people mistake AI-generated language for language created by humans.

In a series of experiments, we analyzed how humans detect AI-generated language in one of the most personal and consequential forms of speech—verbal self-presentation. Self-presentation refers to behaviors designed to control impressions of the self by others (9), while *verbal* self-presentation focuses on the words used to accomplish impression management. In this work, we operationalize self-presentation as self-descriptions of the type prevalent in online profiles (10), e.g., on professional or dating platforms. Researchers have extensively studied the importance of online self-presentation (11–13), showing that impression formation based on self-descriptions is crucial for establishing the trust required for various social interactions (14, 15). AI systems that generate human-like self-presentations may invalidate signals that people rely on when assessing others (16), such as tone or compositional skill. Earlier work on AI-mediated communication (16) has shown that interpersonal trust declines when people suspect that others are using AI systems to generate or optimize their self-presentation (17).

Previous studies suggest that people struggle to discern AI-generated language in different settings (18–20). Here, we go beyond prior work by providing strong evidence that people use flawed heuristics to detect AI-generated language. Using qualitative, quantitative, and computational methods, we reconstruct a set of potential heuristics that people may rely on to detect AI-generated language, expanding on related analyses in previous work (18). We then measure the extent to which people actually use these heuristics and whether the heuristics help or hinder their attempts to distinguish between human- and AI-generated language. Finally, we demonstrate that AI systems can *predict* and *manipulate* whether people perceive AI-generated language as human.

## Results

To examine how people detect AI-generated self-presentations, we performed six experiments broadly patterned after the Turing test (21). While participants in the original test were asked to identify a language-generating machine through a text-based conversation, participants in our studies were asked to judge whether a personal self-presentation was written by a person or generated by an AI system. We trained multiple customized versions of state-of-the-art AI language models (1, 2, 4) to generate self-presentations in three social contexts where trust in a self-presentation is important for decision-making: professional (e.g., job applications) (22), romantic (e.g., online dating) (12), and hospitality services (e.g., Airbnb host profiles) (15). Across three main and three validation experiments, we asked 4,600 participants to read through a total of 7,600 self-presentations—some AI-generated and some collected from real-world online platforms—and indicate which ones they thought were generated by AI. We start by computing the accuracy rates for participants' ability to distinguish between human and AI-generated self-presentations. In our three main experiments, using two different language models to generate verbal self-presentations across three social contexts, participants identified the source of a self-presentation with only 50 to 52% accuracy. These results, with a breakdown by experiments and treatments, are shown in Fig. 1. In the hospitality context (shown in the Left panel), participants correctly identified the source of a self-presentation 52.2% of the time. In the dating context, we introduced experimental treatments testing whether incentivizing participants to increase their efforts (23) would increase their accuracy. In the professional context, we tested whether providing training (18) in the form of feedback would improve participants' judgments. However, participants' accuracy remained close to chance even when offered monetary incentives for accurate assessments (right bar in the second panel in Fig. 1, 51.6%) and when receiving immediate feedback on their evaluations (right bar in the third panel, 51.2%). Further analyses (included in SI Appendix) revealed that no demographic group performed better than others.

Participants' evaluations were not random, however. The observed agreement between participants' judgments was significantly higher than chance (Fleiss' kappa = 0.07, $P < 0.0001$). As the observed accuracy was close to chance, the agreement in participants' assessments must have been due to shared but flawed heuristics that participants relied on to identify AI-generated language. To investigate participants' heuristics for AI-generated language, we next conducted a qualitative analysis of the heuristics participants thought they relied on.

After completing half of the ratings, we asked participants to explain one of their judgments. Two researchers independently coded a sample of their responses and grouped them into themes: content, grammar, tone, and form. These themes are extending categories identified in previous research (18). Participants commonly referred to the content of a self-presentation (40% of responses): Self-presentations with specific content related to family and life experiences led many to infer a human author. Participants also referred to grammatical cues (28%), where first-person pronouns and the mastery of grammar were seen as indicative of language created by humans. Replicating findings from earlier research (18), grammatical errors were associated with a subpar AI by some participants but with fallible human authors by others. Participants also judged the self-presentation source by its tone (24%), associating warm and genuine language with humanity and impersonal, monotonous style with AI-generated language. Details on participants' self-reported explanations of their judgments are included in the SI Appendix.

As self-reports on mental processes can be unreliable and even misleading (24), we conducted additional analyses to evaluate participants' judgments *independently* of their self-reported explanations. While participants may not always know why they did something (25), a multiparadigm approach (26) based on a statistical analysis of their judgments combined with a computational analysis of language features present in the self-descriptions allows us to independently reconstruct heuristics they rely on (27). Rather than drawing conclusions from participants' self-reported heuristics like previous research (18), we used their self-reports as a starting point for extracting potentially relevant language features from the self-presentation texts. We computationally created a range of language features present in the self-presentations, including measurements for personality, sentiment, and perspective (28, 29). We also conducted an additional labeling task to create language features that could not be reliably computed.

For the feature labeling task, we recruited a separate sample of 1,300 crowdworkers. We asked them to read 12 self-presentations and indicate whether they were nonsensical, had grammatical issues, or seemed repetitive. Two to three crowdworkers ($M = 2.3$) evaluated each of the 7,000 human-written and AI-generated self-presentations used in the main experiments. The results indicate that crowdworkers' ratings in the labeling task, to some extent, differentiated between human-written and AI-generated self-presentations. Crowdworkers rated AI-generated self-presentations as nonsensical more often than human-written self-presentations (13.6% vs. 9.6%, $P < 0.0001$). They also rated AI-generated self-presentations as more repetitive (12.7% vs. 7.1%, $P < 0.0001$) and found fewer grammatical issues with AI-generated self-presentations than with human-written self-presentations (14.8% vs. 19.6%, $P <$ 0.0001). These rates differed somewhat between contexts (SI Appendix). We explored the potential of the rating task labels to distinguish AI- and human-written self-presentations. We created a classifier that predicted that a profile was generated when at least one in three raters in the labeling task marked it as nonsensical or repetitive. The classifier predicted the source of a self-presentation with 58.8% accuracy, compared to the 51.7% accuracy participants achieved in the main experiment when directly asked about the source of the self-presentations.

With the language features we created—both computationally and through the labeling task—we quantitatively tested whether the presence of these features was associated with participants' judgments in the main experiments. After a feature selection process, we fit a regression model correlating selected features with participants' perception that a self-presentation was generated by AI. We fit a second model to understand whether the same features are indeed predictive of AI-generated self-presentations. The results suggest that participants relied on several cues in their ratings, some valid and others flawed. Table 1 shows which features were predictive of self-presentations being perceived as AI-generated (on the left) and which features were actually predictive of AI-generated self-presentations (on the right) in the three main experiments.

Some heuristics participants relied on to identify AI-generated self-presentations were indeed indicative of such language. For example, the odds ratios in the top row in Table 1 indicate that self-presentations containing nonsensical content were 10.5% more likely to be seen as AI-generated (left) and, indeed, were 23% more likely to be generated by AI (right). Similarly, self-presentations with repetitive content were 8% more likely to be rated as AI-generated and 47% more likely to be AI-generated in our experiments. However, most heuristics participants relied on were *flawed*: Participants were 5% more likely to rate self-presentations with grammatical issues as AI-generated, although grammatically flawed self-presentations were, in fact, 15% *less* likely to be AI-generated. Participants often rated self-presentations with long words or rare bigrams as generated by AI, while most self-presentations with long words or rare bigrams had been written by humans. Participants also judged first-person speech and family content as more human. However, these cues were not significantly associated with either AI or human-written language. Similarly, self-presentations that were longer, included authentic or spontaneous words (30), or were focused on past events were more likely to be rated as human by participants. However, these features were not significantly associated with human-written or AI-generated self-presentations in our data. Following the correlation analysis, we tested whether the presence of language features in a self-presentation could predict participants' judgments. A regression model based on the features above predicted participants' judgments with 57.6% accuracy when evaluated on a hold-out data set. We also tested whether AI language models can learn to predict human impressions of AI-generated language without feature engineering input from the research team. A current language model (31) with a sequence classification head predicted participants' assessments of AI-generated language with 58.1% accuracy when evaluated on hold-out validation data. These results suggest that the flawed heuristics people rely on to detect AI-generated language allow AI systems to predict their judgments, at least to some extent.

We conducted three additional experiments to validate and extend these findings: If the three main experiments correctly identified features people associate with self-descriptions that are written by humans, self-presentations selected based on the presence of these features would be more likely to be perceived as human-written in independent validation experiments. The validation studies thus tested whether language models can exploit people's flawed heuristics to produce self-presentations perceived as "more human than human." For these validation experiments, we created an additional sample of human-written and AI-generated self-presentations; and used the classifiers trained on participants' judgments in the main studies to create a set of AI-generated self-presentations optimized for perceived humanity. The Fig. 2. shows that participants evaluated the AI-generated self-presentations optimized for perceived humanity as more human than the human-written and the nonoptimized AI-generated self-presentations. Across all three validation experiments (aggregated in the panel on the Right), optimized self-presentations were rated as human more often than regular generated self-presentations (65.7% vs. 51.6%, $P < 0.0001$). The optimized self-presentations were also more likely to be seen as human than self-presentations that were actually written by humans (65.7% vs. 51.7%, $P < 0.0001$). When creating the optimized self-presentations, we used different classifiers in each context to increase generalizability and to independently validate both the regression and language- model-based classifiers. The increase in perceived humanity of optimized self-presentations was strongest in the professional context, where a combination of the regression- and language-model-based classifiers produced self-presentations that were perceived as human 71% of the time.

**Discussion**

Our results reaffirm that humans are not able to detect verbal self-presentations generated by current AI language models. Across contexts and demographics, and independent of effort and expertise, human discernment of AI-generated self-presentation remained close to chance. These results align with recent work showing that humans struggle to detect AI-generated news, recipes, and poetry (18–20), suggesting that the era of the Turing test may be coming to an end. Our results go beyond earlier efforts by providing an empirically grounded explanation of why people fail to identify AI-generated language. Drawing on the extensive literature on deception detection (32–34)**,** we consider two explanations for people's inability to detect AI-generated self-presentation: First, the language generated by state-of-the-art AI systems may be so similar to human-written language that a lack of reliable cues limits accuracy. Second, people's judgments may be inaccurate because they rely on flawed heuristics to detect AI-generated language.

The results of a separate labeling task we conducted suggest that the AI-generated self-presentations in our studies had certain features that people, *in principle,* may be able to detect. The participants in the labeling task rated AI-generated self-presentations as nonsensical and repetitive significantly more often than human-written self-presentations. This finding contradicts the idea that AI-generated language has become entirely indistinguishable from human-written language: While future generations of AI language technologies may change this, the language generated by AI technologies available at the time of the study had some human-detectable features. Yet, when we directly asked participants whether self-presentations were AI-generated in the main experiments—rather than asking them whether self-presentations were nonsensical or repetitive—the accuracy of their judgments remained close to chance.

Our analysis of the heuristics people used to identify AI-generated language provides a more nuanced picture than previous research: While people can sometimes identify certain characteristics of AI-generated language, they rely on other flawed cues that simultaneously impair their judgment. Participants in our studies relied to some extent on functional cues, such as nonsensical and repetitive text, to identify AI-generated self-presentations. Had participants relied on those cues only, they could have achieved a detection accuracy of 58.8%. However, participants also relied on cues like grammatical issues, rare bigrams, or long words to identify AI-generated language, although those cues were more indicative of human-written language in our data. Most other language features that participants relied on to identify human-written language, such as family words or first-person pronouns, were equally present in human-written and AI-generated self-presentations. These misleading heuristics reduced people's accuracy in detecting AI-generated self-presentations to chance, partially explaining why people in our research and in previous work failed to identify AI-generated language (18–20).

People's reliance on flawed intuitive heuristics to detect AI-generated language demonstrates that the increased human-likeness of AI-generated text is *not* necessarily indicative of increased machine intelligence. For example, emphasizing family topics does not require advances in machine intelligence but does increase the perceived humanity of AI-generated self-presentations. Recent work by Ippolito et al. (35) suggests that language model–decoding methods have been optimized for fooling humans at the cost of introducing statistical anomalies easily detected by machines. Previous research also suggests that domain expertise may be somewhat more effective than personal intuition in identifying AI-generated content (23). Rather than interpreting human inability to detect AI-generated language as an indication of machine intelligence, we propose to view it as a sign of human vulnerability. People are unprepared for their encounters with language-generating AI technologies, and the heuristics developed through media exposure and other social contexts are dysfunctional when applied to state-of-the-art AI language systems.

People's inability to detect AI-generated language has important consequences: As demonstrated in the three validation experiments, AI systems can use people's flawed heuristics to manipulate their judgments and produce language perceived as "more human than human." Previous work has shown that not only are people more likely to disclose private information to and adhere to recommendations by nonhuman entities that they perceive as human (36) but they may start distrusting those they believe are using AI-generated language in their communication (17). People's heuristics also can be exploited by malevolent actors. From automated impersonation (8) to targeted disinformation campaigns (37, 38), AI systems could be optimized to undermine human intuition, exacerbating concerns about novel automatized forms of deception, fraud, and identity theft (3–8). Further, new widely available applications like ChatGPT allow anyone to generate human-like text tailored to certain tasks in any requested style (e.g., informal), lowering the barrier to automatically creating language that is deceptively human.

Widespread AI education and technical tools that assist identification (39–41) might improve people's ability to detect AI-generated language to some extent. However, the potential for improving human intuition for the detection of AI-generated language is likely limited (18), and future adaptations of language models may invalidate learned heuristics (35). At the same time, how to transparently identify the use of AI systems in communication is an open and challenging problem. A recent blueprint for an AI Bill of Rights from the US White House calls for "Notice and Explanation" when "an automated system is being used" (42). Similarly, a regulation proposal issued by the EU states that "if an AI system is used to generate or manipulate image, audio or video content that appreciably resembles authentic content, there should be an obligation to disclose that the content is generated through automated means" (43). However, such policies can be difficult to apply in AI-mediated communication (16) where AI technologies modify, augment, or generate communication between people. For example, it hardly seems necessary to add notice to every message people write with AI-enabled autocorrections, smart replies, or translations. Research also shows that typical notice and consent disclosures are largely ignored by users (44).

Identifying context-appropriate and effective disclosure mechanisms for the use of AI in communication is an urgent question that requires further research (45). Our results suggest that one could develop AI language technologies that are *self-disclosing by design*: Rather than training AI language systems to imitate human language, AI systems could be optimized to fulfill their specific communicative function while *preserving the validity of intuitive human judgment* (35). Many AI applications could use language that is clearly *not* written by humans without loss of functionality. AI language systems could reduce the risks of false identifications by producing language that humans intuitively connect to AI sources and avoiding language that people wrongly associate with humanity, such as informal and colloquial speech. Disclosures that preserve the fluidity of communication and support human goals in communication could also be achieved through dedicated *AI accents:* Requiring AI systems to generate language with a dedicated dialect or accent could facilitate people's intuitive judgments without interrupting the flow of communication. Rather than undermining human cognition, AI language technologies that, by design, accommodate the limits and flaws of human judgment may genuinely support human communication and reduce the risk of misuse.

## Materials and Methods

*Experiment design*

The six experiments combined elements of a simplified Turing test (21) with a data labeling task. After providing informed consent, participants were introduced to the *hospitality*, *dating,* or *professional* scenario. They were told that they were browsing an online platform where some users had written their self-presentations while an AI system generated other self-presentations. Participants completed two comprehension checks and rated 16 self-presentations, half generated by a state-of-the-art AI language model. They were asked to evaluate whether each self-presentation was generated by AI on a five-point Likert scale from "definitely AI-generated" to "definitely human-written." Mirroring truth default behaviors observed in deception research (34), participants marked the self-presentations as "likely human-written" or "definitely human-written" in 53.8% of cases. In the remaining 46.2% of cases, they showed suspicion and selected either "not sure," "likely AI-generated," or "definitely AI-generated." To allow for concise analysis, we used these two roughly balanced groups to create a binary signal corresponding to participants' suspicion that a self-presentation may not be human-written. A robustness check using the full scale as the primary outcome metric showed similar results. Halfway through the rating task, participants in the three main experiments were asked to explain their judgment in an open-ended response. Asking participants to explain their reasoning did not change the accuracy of their subsequent ratings (see SI Appendix for details). Following the rating task, participants provided demographic information and indicated their experience with computer programming and AI technologies. Participants were debriefed about their performance and the purpose of the study. The Cornell University Institutional Review Board approved the study protocols. We preregistered the final two validation experiments prior to data collection (https://aspredicted.org/bz7x7.pdf).

We performed the experiments in three social contexts to increase robustness and generalizability. In addition, minor variations across experiments explored auxiliary hypotheses. We used longer self-presentations in the dating- and professional-context experiments to test whether the length of self-presentations limited participants' accuracy. To keep the three main experiments' duration comparable, we reduced the number of rated self-presentations to 12 in these two experiments. To explore the effect of increased effort (23), we offered half of the participants in the *dating* context a bonus payment if they rated at least 75% of the self-presentations correctly. There was no difference in performance between the bonus and no-bonus groups. Finally, to test whether participants could learn to detect generated self-presentations if they received feedback (18), half of the participants in the *professional* context were told whether their choice was correct after every rating, again with no difference in outcomes. An overview of the experimental designs is included in SI Appendix.

*Collecting and generating self-presentations*

We collected data from real-world platforms in each of the three social contexts for the experiments. The data collected were used in two ways: A subset was shown to participants in the experiments, and the full data were used to train state-of-the-art large language models to generate self-presentations. We employed different AI models for generating self-presentations as new and more powerful models became available over the course of this research, providing further generalizability of our findings. An overview of the models used and the setup of each experiment is included in SI Appendix.

For the main experiment in the *hospitality* context, we collected 28,890 verbal self-presentations that contained at least 30 and no more than 60 words from host profiles on Airbnb.com. We drew a random sample of 1,500 human-written self-presentations for the experiment. We fine-tuned a 774M-parameter version of GPT-2 (31) for four epochs with a learning rate of 0.00002 on the collected data. We used the fine-tuned model and nucleus sampling (46) at P = 0.95 to produce 1,500 AI-generated *hospitality* self-presentations. In the *professional* context, we collected 37,450 profile self-presentations with at least 60 and no more than 90 words from Guru.com, a platform where companies find freelance workers for commissioned work. In the *dating* context, we used a publicly available dataset of 59,940 OkCupid.com self-presentation essays collected with the platform operators' permission (47). We drew a random sample of 1,000 human-written self-presentations for the *professional* and *dating* main experiments. We used the full set of collected self-presentations in each of these contexts to fine-tune a 13B-parameter version of GPT-3 (1) for four epochs with a learning rate multiplier of 0.1. We used these fine-tuned models to produce 1,000 AI-generated self-presentations for each experiment with temperature sampling at t = 0.9.

We confirmed that there were no duplicate self-presentations and used multiple techniques to check that the models did not plagiarize the training data. For example, we searched for identical sentences in the training data and AI-generated text and found that 95% of sentences in the AI-generated texts were not present in the training data. As we found no signs of substantial plagiarism, we used the AI-generated self-presentations without further preprocessing.

*Predicting responses and optimizing self-presentations.*

We developed a set of text-based language features for the quantitative language analysis of participants' judgments in the three main experiments. The full set of about 180 features is included in SI Appendix. We used two approaches to create these features: One set of language features were computational features that could be automatically extracted from the text. For the computational features, we manually developed measures motivated by participants' explanations of their judgments. To this initial set, we added readability scores, emotion language classification, and other psychological language features (29). We relied on a labeling task for features that could not be reliably computed. We created three additional key features by recruiting crowdworkers (N = 1,300) to label which self-presentations seemed nonsensical, contained repetitive text, or had grammatical issues.

For the prediction task, to reduce overfitting and increase interpretability, we reduced the set of relevant features to 15 in a feature selection process based on lasso regression performed on 20% of the self-presentations. Table 1 reports the coefficients of a logistic regression model fitted to 4,900 self-presentations (70%) that were not used for feature selection. In addition, to test whether modern language models can learn to predict human perceptions of AI-generated language without predeveloped features, we trained a large language model with a sequence classification head on 4,900 self-presentations to predict participants' judgments. We trained the 117M parameter version of GPT-2 (31) with a learning rate of 0.00005 on 70% of the data and stopped training when performance on the validation data set (20%) decreased. The predictive accuracy of the regression and sequence classification models was evaluated on a separate hold-out data set consisting of the 700 remaining self-presentations (10%).

*Generating language optimized for perceived humanity*

For the three validation experiments, we drew a separate sample of 100 human-written self-presentations from the collected data. We created an additional set of 100 AI-generated self-presentations using the methods described in the main studies. We then produced an additional set of 100 self-presentations optimized for perceived humanity. To create these optimized self-presentations, we first generated a large number of self-presentations in each context using the same models as in the initial experiments. We then used the classifiers developed above to select self-presentations that the model predicted would be perceived as written by humans. We employed different classifiers to select self-presentations in each context to increase generalizability and validate both the regression and the language-model-based classifier. In the *dating* context, we used the regression-based classifier on the GPT-3 output to select those generated self-presentations that were more likely to be perceived as human-written. In the *hospitality* context, we used a classifier based on language models to perform the same task, connecting the GPT-2 generation model with the GPT-2 sequence classifier trained to predict participants' evaluation of self-presentations. In the *professional* context, we combined the regression and language-model classifiers using an ensemble approach. In each context, we selected the top 20% percentile of self-presentations that the classifier predicted were likely to be perceived as human-written. We drew a random sample of 100 self-presentations optimized for perceived humanity from these sets for each of the three validation experiments.

*Participant recruitment*
For the main experiment in the *hospitality* context, we recruited a US-representative sample of 2,000 participants through Lucid (48). The experiment's results indicated that participants' answers did not vary significantly across demographics and that a smaller sample size would be sufficient for follow-up experiments. In the main *dating* and *professional* experiments, we recruited two gender-balanced samples of 1,000 US-based participants each from Prolific (49), a platform that enabled us to process bonus payments. Participants from Prolific had a median age of 37 y, 67% had a college degree, and 27% were at least somewhat familiar with computer programming. The median time participants spent evaluating each self-presentation was 14.3 s (mean = 23.1, SD = 39.6). In return for their time, participants received compensation of $1.40 at a rate of about $12.5 per hour. Participants in the bonus condition in the *dating* context received an additional $3 bonus payment if they correctly rated at least 9 out of 12 self-presentations. We recruited a separate set of 1,300 crowdworkers to create the language features that could not be reliably computed for the 7,000 self-presentations in the main experiments. These crowdworkers were recruited from the same platforms as the participants in the main experiments and rated 12 self-presentations each, receiving compensation of $1.10. We recruited 200 participants for each of the three validation experiments on the respective platforms. Tasks and payments were analogous to the main experiments.

*Limitations and ethics statement*
Our results are limited to the current generation of language models and people's current heuristics for AI-generated language. Developments in technology and culture may change both the heuristics people rely on and the characteristics of AI-generated language. However, it is unlikely that in other cultural settings or for future generations of language models, human intuition will naturally coincide with the characteristics of AI-generated language. Our findings show that humans' flawed heuristics leave them vulnerable to large-scale automated deception. In disclosing this vulnerability, we face ethical tensions similar to cybersecurity researchers: On the one hand, publicizing a vulnerability increases the chance that someone will exploit it; on the other, only through public awareness and discourse effective preventive measures can be taken at the policy and development level. While risky, decisions to share vulnerabilities have led to positive developments in computer safety (50).

**Acknowledgments**

We thank Benjamin Kim Carson for his assistance in collecting the self-presentation data, evaluating the qualitative data, and developing the language features.

## Data sharing

The data and code for the analyses performed across the three main studies and three validation experiments are publicly available through an Open Science Foundation repository (https://osf.io/284yv/). Previously published data were used for this work (47).

## Figures and Tables

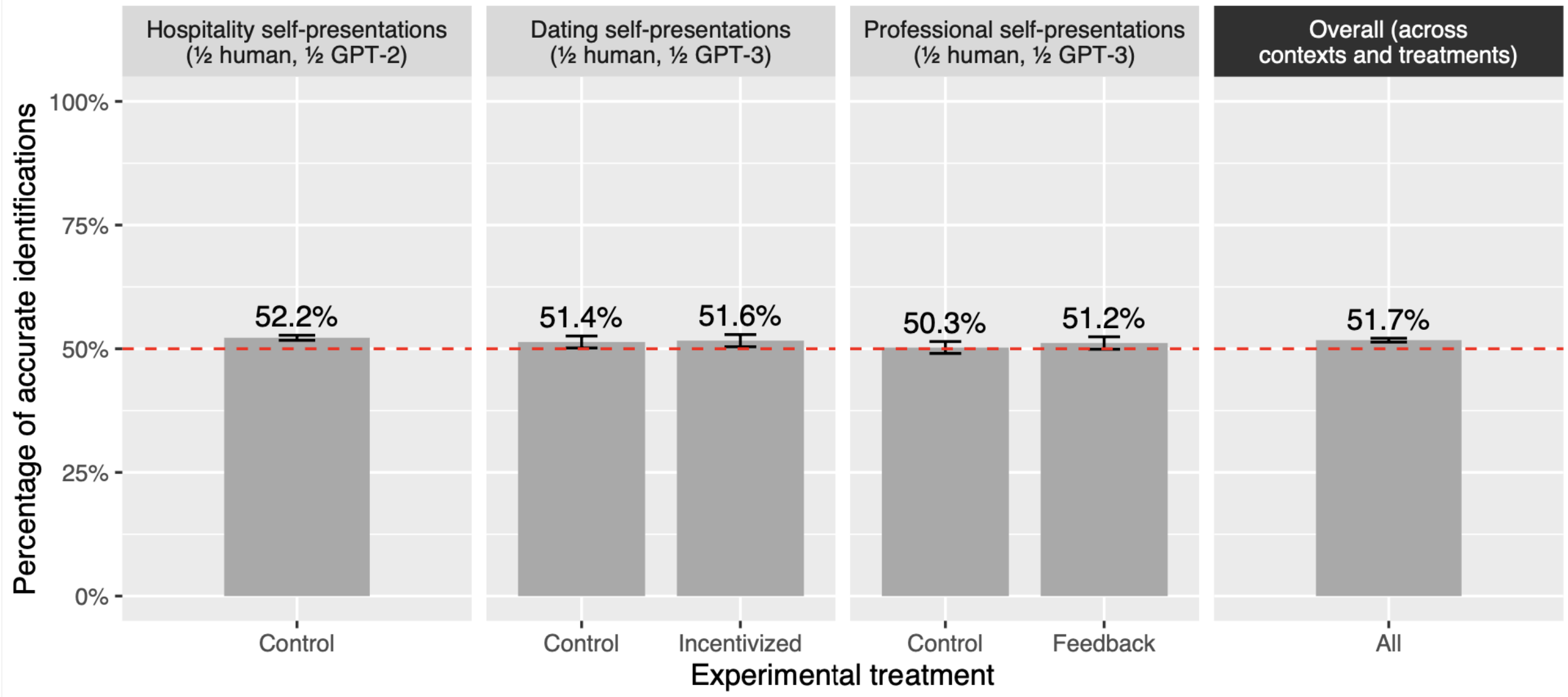

***Figure 1.*** *Participants could not detect self-presentations generated by the current AI language models beyond chance in the three main experiments. Error bars represent 95% CIs for 6,000 to 16,000 judgments of 2,000 to 3,000 self-presentations per bar. Across three social contexts, discernment remained close to chance. Providing monetary incentives for accurate answers or telling participants whether their answers were correct did not increase accuracy.*

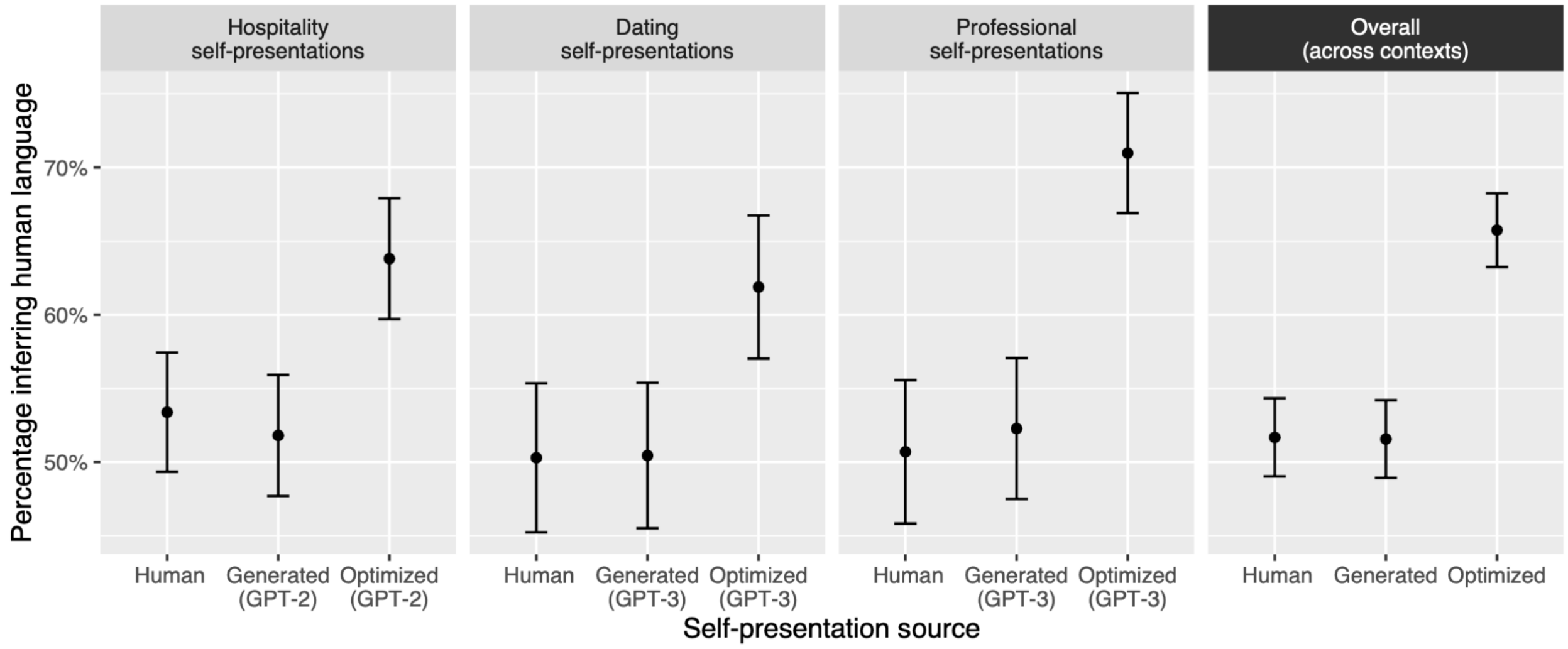

***Figure 2.*** *The three validation experiments show that AI systems can exploit humans' flawed heuristics to generate optimized self-presentations (rightmost in each pane) more likely to be perceived as human than human-written self-presentations (leftmost) and regular AI-generated profiles (center). Error bars represent 95% CIs for 350 to 450 judgments of 100 self-presentations per bar.*

***Table 1:*** *Logistic regression models predicting (1) whether participants in the three main experiments rated a self-presentation as AI-generated and (2) whether a self-presentation was actually generated by AI. Only nonsense, repetition, and conversational words were functional cues (top section), indicated by aligned odds ratios in models (1) and (2). The remaining features indicative of participants' heuristics were either inversely related (center) or unrelated (bottom) to features indicative of the actual source of the self-presentation.*

| | *Dependent variable:* | |
|---|---|---|
| | (1) Perceived as AI-generated (odds ratios with 95% CI) | (2) Actually AI-generated (odds ratios 95% CI) |
| Aligned features | | |
| Nonsensical content † | 1.105*** (1.085, 1.126) | 1.233*** (1.169, 1.296) |
| Repetitive content † | 1.083*** (1.059, 1.106) | 1.470*** (1.379, 1.561) |
| Conversational words | 0.947*** (0.925, 0.970) | 0.898** (0.829, 0.967) |
| Misaligned features | | |
| Grammatical issues † | 1.048*** (1.028, 1.069) | 0.851*** (0.788, 0.913) |
| Rare bigrams | 1.042*** (1.019, 1.065) | 0.666*** (0.596, 0.736) |
| Long words | 1.034** (1.009, 1.059) | 0.783*** (0.706, 0.861) |
| Contractions | 0.947*** (0.924, 0.970) | 1.134*** (1.065, 1.203) |
| Nonindicative | | |
| Second-person pronouns | 1.059*** (1.038, 1.079) | 0.970 (0.908, 1.032) |
| Filler words | 1.009 (0.990, 1.027) | 1.119* (1.021, 1.218) |
| Swear words | 0.969** (0.948, 0.989) | 0.965 (0.905, 1.024) |
| Authentic words | 0.946*** (0.921, 0.971) | 0.945 (0.870, 1.021) |
| Focus on past | 0.938*** (0.917, 0.959) | 1.002 (0.940, 1.064) |
| First-person pronouns | 0.925*** (0.886, 0.963) | 0.992 (0.868, 1.117) |
| Family words | 0.910*** (0.889, 0.932) | 1.014 (0.950, 1.077) |
| Word count | 0.904*** (0.874, 0.935) | 1.076 (0.986, 1.165) |
| *Constant* | 0.850*** (0.830, 0.870) | 1.007 (0.947, 1.068) |
| Observations | 38,866 | 4,690 |
| Log Likelihood | -26,318.460 | -3,029.542 |
| Akaike Inf. Crit. | 52,670.930 | 6,093.085 |
| *Note:* | †manually labeled feature, *p**p***p<0.001 | |

## Supporting Information

Below we provide additional information on several aspects of our experiments. Table S1 summarizes the treatment, stimuli, and recruitment methods used across the six studies and three labeling tasks. Table S2 shows a sample of self-presentations for each study and treatment group.

Table S3 shows the results of an auxiliary analysis testing whether certain groups are better at detecting AI-generated language than others. Older participants were slightly more likely to detect AI-generated self-presentations, with participants older than 50 achieving an accuracy of 53% (compared to 51% for younger participants). No gender or ethnic group performed better than others. Participants with a university degree performed about 1% worse than those without, and self-reported technical knowledge was not correlated with more accurate ratings. Neither the time taken for the judgment nor the length of profiles predicted higher judgment accuracy. Across contexts, groups, and treatments, participants could not detect AI-generated self-presentations.

Table S4 and S5 provide further detail on the qualitative analysis of participants' explanations of why they thought certain self-presentations were AI-generated or human-written. Two researchers independently coded a sample of responses into themes to provide an overview of participants' self-reported heuristics. Table S4 presents an overview of recurring themes. Participants most commonly referred to the content of a self-presentation (blue-shaded regions in Table S4 representing 40% of responses). The participants reported associating specific content related to family and life experiences with language written by humans and generic or nonsensical content with AI-generated language. Participants also reported basing their decisions on grammatical cues (gray, 28%), where first-person pronouns and the mastery of grammar were mentioned as indicative of human-generated language. Some participants saw grammatical errors as associated with a subpar AI, but others claimed they associated them with fallible human authors. Another category of cues mentioned by participants was the tone (green, 24%). Participants reported associating warm and genuine language with humanity and impersonal, monotonous style with AI-generated language. The codebook, theme frequencies, and sample responses are shown in Table S5. Table S6 provides a complete overview of the developed language features and statistical summaries.

Prior research suggests that asking participants to explain their responses could have changed their subsequent evaluations or degraded performance (1,2). We thus conducted an analysis testing whether participants' performance had changed after being asked to explain their judgment. The results are shown in Figure S1. There was no evidence for such change in our data as participants' accuracy before and after the open-ended response did not change across any of the three contexts. Note that open-ended responses were only solicited for the three main experiments. The validation experiments did not include open-ended responses, showing similar outcomes and providing further evidence that participants' ratings (and our findings) were not affected by the explanations.

Figure S2 shows how crowdworkers evaluated human-written and AI-generated self-presentations in a separate labeling task when asked whether the text was nonsensical, seemed repetitive, or had grammatical issues. Crowdworkers were significantly more likely to rate AI-generated self-presentations as nonsensical (13.6% vs. 9.6%, $p<0.0001$). This was the case in the hospitality context, in particular, where we had used the older GPT-2 model to generate self-presentations. Crowdworkers also rated generated self-presentations as more repetitive (12.7% vs. 7.1%%, p<0.0001), particularly in the professional context. Finally, crowdworkers labeled generated self-presentations as having fewer grammatical issues than human-written text (14.8% vs. 19.6%, p<0.0001). This difference was most pronounced in the dating and professional contexts where we had used the more advanced GPT-3 model to generate self-presentations.

**SI References**

***Table S1:*** *Overview of experiments*

| Context | Stimuli | Treatment | Recruitment |
|---|---|---|---|
| Main study 1: Hospitality | 1,500 self-presentations from Airbnb and 1,500 generated by GPT-2; 30-60 words each; 16 per subject | Within-subject variation of self-presentation type | N = 2,000 US-representative sample via Lucid |
| Main study 2: Dating | 1,000 self-presentations from OkCupid and 1,000 generated by GPT-3; 60-90 words; 12 per subject | Within-subject variation of self-presentation type and between-subject bonus payments for correct ratings | N = 1,000 gender-balanced sample via Prolific |
| Main study 3: Professional | 1,000 self-presentations from Guru and 1,000 generated by GPT-3; 60-90 words each; 12 per subject | Within-subject variation of self-presentation type and between-subject feedback on answers | N = 1,000 gender-balanced sample via Prolific |
| Validation study 1: Hospitality | 100 self-presentations from Airbnb, 100 generated by GPT-2, and 100 optimized using the language model classifier; 16 per subject | Within-subject variation of self-presentation type | N = 250 US-representative sample via Lucid |
| Validation study 2: Dating | 100 self-presentations from OkCupid, 100 generated by GPT-3, and 100 optimized by the regression classifier; 16 per subject | Within-subject variation of self-presentation type | N = 200 gender-balanced sample via Prolific |
| Validation study 3: Professional | 100 self-presentations from Guru, 100 generated by GPT-3, and 100 optimized using an ensemble classifier; 16 per | Within-subject variation of self-presentation type | N = 200 gender-balanced sample via Prolific |
| | subject | | |
| Labeling task 1: Hospitality | 1,500 self-presentations from Airbnb and 1,500 generated by GPT-2; 30-60 words each; 12-16 per crowdworker | None | N = 600 US-representative sample via Lucid |
| Labeling task 2: Dating | 1,000 self-presentations from OkCupid and 1,000 generated by GPT-3; 60-90 words; 12 per crowdworker | None | N = 350 gender-balanced sample via Prolific |
| Labeling task 3: Professional | 1,000 self-presentations from Guru and 1,000 generated by GPT-3; 60-90 words each; 12 per crowdworker | None | N = 350 gender-balanced sample via Prolific |

***Table S2:*** *Self-presentation examples*

| Context | Source | Example |
|---|---|---|
| Hospitality | Human | My family has lived in DC for the past several years. Some of our favorite things about living on Capitol Hill are running through the neighborhood, exploring all the museums and exhibits that are walking distance from our home, and having a variety of great food offerings only steps away. |
| Hospitality | Generated (GPT-2) | A teacher and young entrepreneur, I love to ski and travel. My wife & I have lived in Vermont for the past 10 years and love the beauty and the snow that we get to ski during the summer. |
| Hospitality | Generated (GPT-2) & optimized (regression) | My husband and I have lived in Denver for 20 years. A few summers ago we visited my two brothers who live elsewhere so we decided to make our home available for others to enjoy as well. We love traveling in Europe, South America and anywhere new! Welcome to your home away from home. |
| Dating | Human | i'm an elementary school social worker and find my job both fulfilling and frustrating. an la native, i've also lived in the midwest and new england. i've been in sf for about 6 years now and love the people, politics, and food here. but, i do miss having seasons and look forward to my annual vacations back in the midwest, which generally involve lounging on a lake and drinking bell's beer. i enjoy being fit, active, and healthy, though i do eat ice cream for dinner on occasion. |
| Dating | Generated (GPT-3) | i just moved to the city last august and really don't know many people here yet. i'm interested in hanging out and maybe even finding someone special. i would love to be able to spend time together without any drama and want to get to know each other better. i'd love to find someone that i can share all of these exciting things in life with like art galleries, theatre, dinner, etc... |
| Dating | Generated (GPT-3) & optimized (GPT-2) | hey i moved to sf about 2 years ago, it's such a great city..i like to explore the city, always trying to find new hangouts and food... i've travelled a lot around the world and would love to travel more. i'm easy going and down to earth, i know what i want in life and am |
| | | working towards my goals. message me if you want to know more :) |
| Professional | Human | I have 19 years of journalism experience. My work has appeared in daily and weekly newspapers, international trade magazines and textbooks. I also have worked in broadcast news, and my reporting has been picked up by the Associated Press. For six years, my interviews focused on C-level execs at Fortune 500 power companies, tech startups and government. In 2015, I became managing editor of a publication in the petroleum and fluid handling equipment industry. |
| Professional | Generated (GPT-3) | My name is Gary Stauch and I have been in the computer and electronics business for over 30 years. I have a A.S. in electronics, a B.S. in computer science and I am a registered professional engineer in Texas. In addition to my own company, I have worked for several others in the design and deployment of large scale network infrastructure in the data center and enterprise server market. I have designed and developed server platforms, workstations, servers, switches, routers and other devices that are part of large scale networks. |
| Professional | Generated (GPT-3) & optimized (regression and GPT-2) | I am a mother of three and a grandmother of two. I live in beautiful Iowa and have been here all my life. I enjoy doing different things but I am a master at none. I love to tell stories and make people smile with laughter. I am very well at reading people and knowing what to do to get the job done. I am very good at multi-tasking. I am very organized and very well at using my time. |

***Table S3:*** *Regression coefficients predicting the accuracy of a judgment based on treatment, social context, and participant demographics. No group performed much above chance level.*

| | *Dependent variable:* |
|---|---|
| | Likelihood of accurate assessment OR (95% CIs) |
| Context: Dating profiles | 0.974 (0.882, 1.065) |
| Context: Professional profiles | 0.926 (0.845, 1.007) |
| Treatment: Feedback | 1.038 (0.966, 1.110) |
| Treatment: Incentives | 1.022 (0.944, 1.100) |
| Age | **1.002**$^{**}$ (1.001, 1.003) |
| Gender: Female | 1.002 (0.967, 1.036) |
| Gender: Non-binary | 1.010 (0.834, 1.186) |
| Race: African American | 0.959 (0.895, 1.022) |
| Race: Asian | 1.055 (0.976, 1.134) |
| Race: Hispanic | 1.005 (0.940, 1.069) |
| Race: Other | 0.973 (0.887, 1.059) |
| Level of education | **0.986**$^{**}$ (0.976, 0.996) |
| Technical knowledge | 1.006 (0.982, 1.030) |
| Rating: Time taken | 1.000 (1.000, 1.001) |
| Profile: Word count | 1.000 (0.998, 1.002) |
| Constant | 1.045 (0.925, 1.166) |
| Observations | 53,411 |
| Log Likelihood | -37,199.800 |
| Akaike Inf. Crit. | 74,435.610 |
| *Note:* | $^{*}$p$^{**}$p$^{***}$p<0.001 |

***Table S4.*** *Themes in participants' explanations of why they thought a self-presentation was human or generated language. N = 800, tile areas correspond to theme prevalence reported in brackets. Heuristics are classified by whether they refer to the content (blue), tone (green), grammar (gray), or form (red) of a self-presentation. Lighter tiles show cue that were associated with generated language.*

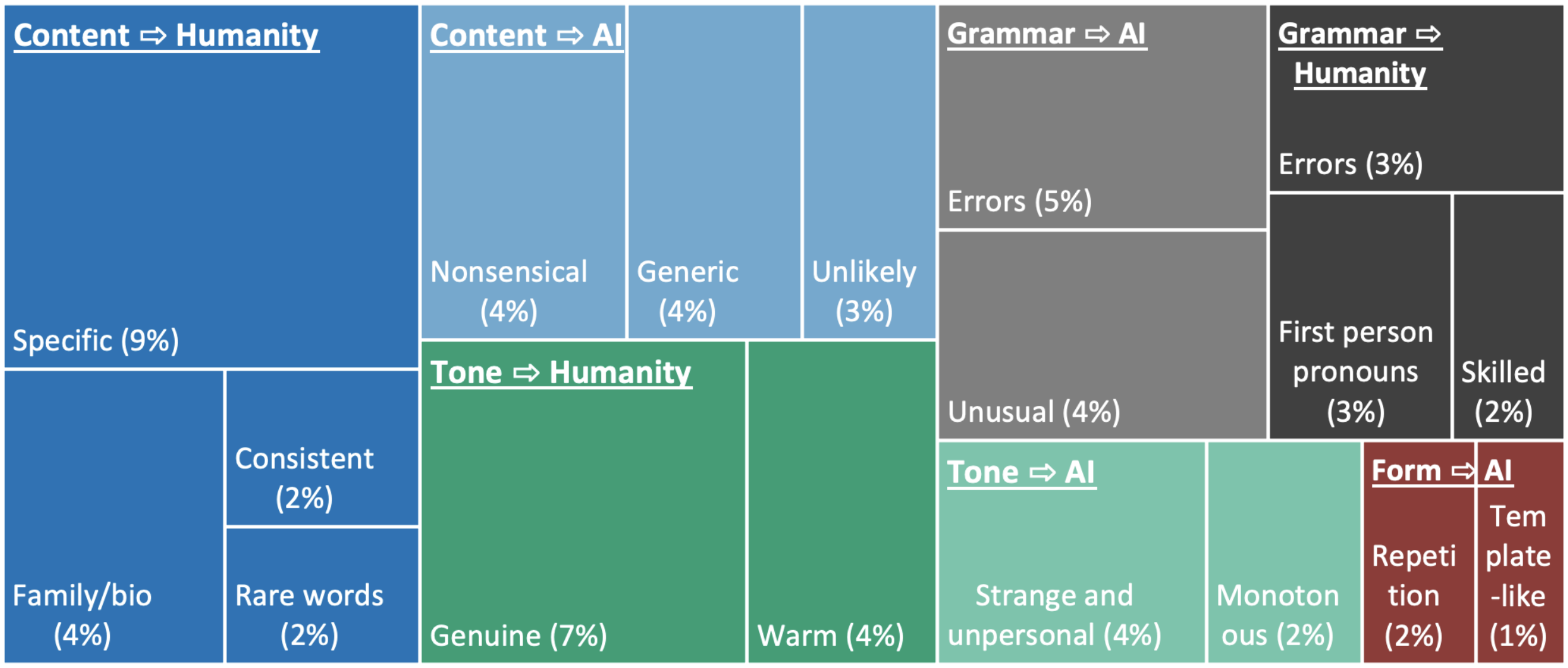

***Table S5:*** *Examples themes and codes in participants' explanations of judgments*

| Category | Code | Freq. | Example |
|---|---|---|---|
| Content cues for AI | Nonsensical content | 7% | "'travel here from around the world' in third sentence doesn't make sense" |
| Content cues for AI | Generic content | 6% | "seems just a bit to generic and a bit random" |
| Content cues for AI | Unlikely content | 4% | "A full time manager at a nuclear plant doesn't travel frequently enough to care about hotel amenities." |
| Content cues for | Specific content | 14% | "How detailed descriptions were" |
| Humanity | | | |
| Content cues for Humanity | Family and biography | 6% | I determine this is a person because he says him and his wife and son travel and go places on there free time” |
| Content cues for Humanity | Consistent | 3% | “Based primarily on the content, and whether each part of the statement made sense logically and thematically with the rest.” |
| Form cues for AI | Repetitive | 2% | “the repetition of the sentences make the whole thing sound lifeless and robotic.” |
| Form cues for AI | Template-like | 2% | “I looked for a stock template response for AI, or for signs of a disjointed copy and paste from real user statements.” |
| Grammar cues for AI | Errors | 7% | “If things are worded incorrectly.” |
| Grammar cues for AI | Unusual punctuation | 7% | “There should be a comma after ‘I’m Kellie’” |
| Grammar cues for Humanity | Errors | 5% | “Believe there was a grammar error where it should have been knowledgeable” |
| Grammar cues for Humanity | 1st person speech | 4% | “Using I, me, we language” |
| Grammar cues for Humanity | Good grammar | 3% | “The English is good, but not great. It possibly is written by someone who is ESL.” |
| Grammar cues for Humanity | Rare words | 3% | “Certain words that were unusual.” |
| Tone cues for AI | Strange and unpersonal | 6% | “The personal touch is very unnatural sounding.” |
| Tone cues for AI | Monotonous | 3% | “most people either put in little or more thought and AI just feels like a perfect monotone read” |
| Tone cues for Humanity | Genuinely personal | 10% | “one can have a few replies per question and then have the AI Place together; but this isnt random.. it is Genuine”” |
| Tone cues for Humanity | Warm and welcoming | 6% | “Its how the phrase comes across, An AI Having Emotion…” |

***Table S7:*** *Overview of language features and their correlations with participants’ judgments.*

| Feature Name | Mean | SD. | Min | Max | Cor. with ratings | Cor. with source |
|---|---|---|---|---|---|---|
| Nonsensical (manual labels) | 0.117 | 0.233 | 0 | 1 | 0.086 | 0.114 |
| Repetitive (manual labels) | 0.099 | 0.222 | 0 | 1 | 0.127 | 0.057 |
| Grammatical issues (manual) | 0.172 | 0.281 | 0 | 1 | -0.086 | 0.057 |
| LIWC Achieve | 2.325 | 2.535 | 0 | 17.72 | 0.037 | -0.009 |
| LIWC Acquire | 0.492 | 0.941 | 0 | 9.72 | -0.012 | 0.01 |
| LIWC Adjective | 6.968 | 3.797 | 0 | 30.95 | 0.029 | -0.007 |
| LIWC Adverb | 3.898 | 3.038 | 0 | 22.22 | -0.094 | 0.023 |
| LIWC Affect | 7.368 | 4.983 | 0 | 34.48 | 0.028 | 0.024 |
| LIWC Affiliation | 3.15 | 4.238 | 0 | 25.81 | 0.033 | 0.032 |
| LIWC Allnone | 0.854 | 1.374 | 0 | 12.9 | 0.017 | 0.001 |
| LIWC Allpunc | 17.037 | 8.73 | 0 | 257.14 | -0.009 | -0.046 |
| LIWC Allure | 9.614 | 4.967 | 0 | 32.35 | -0.056 | 0.09 |
| LIWC Analytic | 56.357 | 27.358 | 1 | 99 | 0.098 | -0.051 |
| LIWC Apostro | 1.75 | 2.249 | 0 | 21.67 | -0.109 | 0.03 |
| LIWC Article | 5.938 | 3.066 | 0 | 20.45 | 0.013 | 0.088 |
| LIWC Assent | 0.035 | 0.259 | 0 | 8.82 | -0.047 | -0.013 |
| LIWC Attention | 0.615 | 1.259 | 0 | 10.64 | 0.003 | 0.019 |
| LIWC Auditory | 0.336 | 0.976 | 0 | 11.54 | -0.011 | -0.036 |
| LIWC Authentic | 72.388 | 30.232 | 1 | 99 | -0.197 | 0.031 |
| LIWC Auxverb | 7.599 | 3.585 | 0 | 25 | -0.077 | 0.112 |
| LIWC Bigwords | 20.104 | 8.673 | 0 | 68.42 | 0.123 | -0.128 |
| LIWC Cause | 0.94 | 1.367 | 0 | 9.52 | 0.033 | -0.014 |
| LIWC Certitude | 0.35 | 0.89 | 0 | 9.3 | -0.038 | -0.01 |
| LIWC Clout | 33.423 | 35.196 | 1 | 99 | 0.162 | 0.01 |
| LIWC Cognition | 8.361 | 5.307 | 0 | 36.67 | -0.012 | -0.011 |
| LIWC Cogproc | 7.457 | 5.005 | 0 | 36.67 | -0.017 | -0.01 |
| LIWC Comm | 1.236 | 1.758 | 0 | 17.65 | -0.035 | -0.011 |
| LIWC Comma | 5.566 | 4.722 | 0 | 42.11 | 0.024 | -0.075 |
| LIWC Conflict | 0.033 | 0.248 | 0 | 5 | -0.02 | -0.019 |
| LIWC Conj | 8.083 | 3.071 | 0 | 25.3 | -0.033 | 0.055 |
| LIWC Conversation | 0.24 | 0.801 | 0 | 21.05 | -0.089 | -0.026 |
| LIWC Culture | 0.988 | 1.961 | 0 | 19.05 | 0.06 | -0.02 |
| LIWC Curiosity | 0.983 | 1.601 | 0 | 12.5 | -0.007 | 0.022 |
| LIWC Death | 0.02 | 0.19 | 0 | 3.61 | -0.007 | -0.012 |
| LIWC Det | 11.627 | 4.017 | 0 | 27.66 | -0.021 | 0.06 |
| LIWC Dic | 88.99 | 6.611 | 36.84 | 100 | -0.092 | 0.164 |
| LIWC Differ | 2.054 | 2.199 | 0 | 14.71 | -0.04 | 0.013 |
| LIWC Discrep | 1.208 | 1.687 | 0 | 12.2 | -0.005 | 0.01 |
| LIWC Drives | 6.244 | 4.802 | 0 | 29.41 | 0.069 | 0.008 |
| LIWC Emo Anger | 0.026 | 0.233 | 0 | 5.88 | -0.022 | -0.023 |
| LIWC Emo Anx | 0.033 | 0.277 | 0 | 8.22 | -0.015 | -0.023 |
| LIWC Emo Neg | 0.132 | 0.56 | 0 | 9.09 | -0.032 | -0.032 |
| LIWC Emo Pos | 2.502 | 2.662 | 0 | 17.65 | -0.012 | 0.033 |
| LIWC Emo Sad | 0.016 | 0.173 | 0 | 5.08 | 0.023 | -0.01 |
| LIWC Emotion | 2.679 | 2.747 | 0 | 20.59 | -0.018 | 0.023 |
| LIWC Ethnicity | 0.122 | 0.675 | 0 | 16.39 | 0.002 | -0.034 |
| LIWC Exclam | 0.76 | 1.68 | 0 | 26.58 | -0.007 | -0.024 |
| LIWC Family | 0.602 | 1.465 | 0 | 12.9 | -0.083 | 0.011 |
| LIWC Fatigue | 0.014 | 0.164 | 0 | 4 | -0.022 | 0.001 |
| LIWC Feeling | 0.267 | 0.738 | 0 | 6.67 | 0.018 | -0.006 |
| LIWC Female | 0.426 | 1.197 | 0 | 19.35 | -0.008 | -0.015 |
| LIWC Filler | 0.005 | 0.098 | 0 | 4.11 | -0.015 | 0.015 |
| LIWC Focusfuture | 0.919 | 1.624 | 0 | 16.67 | 0.022 | -0.001 |
| LIWC Focuspast | 2.345 | 2.636 | 0 | 15.38 | -0.111 | 0.008 |
| LIWC Focuspresent | 5.2 | 2.984 | 0 | 24.14 | 0.003 | 0.072 |
| LIWC Food | 0.737 | 1.657 | 0 | 19.05 | -0.01 | 0.009 |
| LIWC Friend | 0.466 | 1.053 | 0 | 14.29 | 0.025 | 0.039 |
| LIWC Fulfill | 0.153 | 0.527 | 0 | 5.56 | 0.036 | -0.017 |
| LIWC Function | 51.71 | 8.185 | 1.32 | 79.41 | -0.129 | 0.162 |
| LIWC Health | 0.31 | 1.037 | 0 | 17.86 | -0.006 | -0.01 |
| LIWC Home | 0.721 | 1.531 | 0 | 22.86 | 0.021 | -0.007 |
| LIWC I me | 7.962 | 4.525 | 0 | 24.39 | -0.212 | 0.031 |
| LIWC Illness | 0.024 | 0.249 | 0 | 6.25 | 0.019 | -0.026 |
| LIWC Insight | 1.674 | 1.994 | 0 | 15 | 0.022 | -0.029 |
| LIWC Ipron | 2.301 | 2.483 | 0 | 22.06 | -0.005 | 0.029 |
| LIWC Lack | 0.051 | 0.381 | 0 | 6.9 | -0.003 | -0.021 |
| LIWC Leisure | 1.975 | 2.788 | 0 | 19.35 | -0.043 | -0.004 |
| LIWC Lifestyle | 8.156 | 5.838 | 0 | 40 | 0.025 | -0.013 |
| LIWC Linguistic | 66.286 | 9.102 | 6.58 | 91.18 | -0.122 | 0.156 |
| LIWC Male | 0.572 | 1.223 | 0 | 15.69 | -0.004 | 0.009 |
| LIWC Memory | 0.031 | 0.259 | 0 | 4.76 | 0.02 | -0.021 |
| LIWC Mental | 0.022 | 0.246 | 0 | 8 | -0.022 | 0.011 |
| LIWC Money | 1.089 | 2.075 | 0 | 20.51 | 0.065 | -0.012 |
| LIWC Moral | 0.204 | 0.69 | 0 | 8.11 | -0.016 | -0.04 |
| LIWC Motion | 2.131 | 2.293 | 0 | 16.13 | -0.032 | 0.018 |
| LIWC Need | 0.282 | 0.86 | 0 | 8.86 | 0.022 | -0.009 |
| LIWC Negate | 0.521 | 1.082 | 0 | 12.5 | -0.057 | -0.019 |
| LIWC Netspeak | 0.184 | 0.686 | 0 | 21.05 | -0.082 | -0.028 |
| LIWC Nonflu | 0.022 | 0.196 | 0 | 4.35 | -0.022 | 0.003 |
| LIWC Number | 1.364 | 1.965 | 0 | 27.27 | -0.031 | -0.026 |
| LIWC Otherp | 1.901 | 3.802 | 0 | 163.77 | 0.014 | -0.043 |
| LIWC Perception | 11.3 | 5.608 | 0 | 43.24 | -0.021 | 0.016 |
| LIWC Period | 7.001 | 4.89 | 0 | 245.71 | 0.003 | 0.019 |
| LIWC Physical | 1.785 | 2.432 | 0 | 23.81 | -0.022 | -0.006 |
| LIWC Polite | 0.38 | 0.995 | 0 | 10 | 0.043 | -0.044 |
| LIWC Politic | 0.185 | 0.802 | 0 | 13.64 | 0.031 | -0.005 |
| LIWC Power | 0.855 | 1.535 | 0 | 15.66 | 0.073 | -0.054 |
| LIWC Ppron | 11.167 | 4.369 | 0 | 27.91 | -0.133 | 0.064 |
| LIWC Prep | 13.841 | 4.042 | 0 | 29.51 | -0.013 | 0.035 |
| LIWC Pronoun | 13.468 | 5.147 | 0 | 32.65 | -0.115 | 0.068 |
| LIWC Prosocial | 0.887 | 1.507 | 0 | 13.33 | 0.089 | -0.019 |
| LIWC Qmark | 0.061 | 0.431 | 0 | 13.24 | -0.016 | -0.002 |
| LIWC Quantity | 3.614 | 2.857 | 0 | 18.82 | -0.067 | -0.018 |
| LIWC Relig | 0.085 | 0.561 | 0 | 17.65 | -0.001 | -0.029 |
| LIWC Reward | 0.228 | 0.682 | 0 | 6.67 | 0.043 | -0.012 |
| LIWC Risk | 0.094 | 0.432 | 0 | 7.69 | 0.028 | -0.045 |
| LIWC Sexual | 0.026 | 0.246 | 0 | 7.81 | -0.032 | -0.041 |
| LIWC Shehe | 0.131 | 0.767 | 0 | 13.89 | 0.08 | -0.005 |
| LIWC Socbehav | 4.371 | 3.262 | 0 | 23.33 | 0.032 | -0.019 |
| LIWC Social | 11.563 | 6.541 | 0 | 48.72 | 0.074 | 0.028 |
| LIWC Socrefs | 6.542 | 5.332 | 0 | 36.17 | 0.07 | 0.045 |
| LIWC Space | 7.688 | 4.578 | 0 | 30.3 | -0.016 | 0.02 |
| LIWC Substances | 0.084 | 0.465 | 0 | 10.2 | 0.019 | 0.018 |
| LIWC Swear | 0.025 | 0.213 | 0 | 4.23 | -0.058 | -0.02 |
| LIWC Tech | 0.682 | 1.653 | 0 | 19.05 | 0.055 | -0.007 |
| LIWC Tentat | 1.583 | 2.181 | 0 | 15.79 | -0.041 | 0.009 |
| LIWC They | 0.283 | 0.863 | 0 | 10 | 0.035 | 0.024 |
| LIWC Time | 3.959 | 2.977 | 0 | 24.39 | -0.088 | -0.01 |
| LIWC Tone | 79.83 | 26.516 | 1 | 99 | 0 | 0.03 |
| LIWC Tone Neg | 0.318 | 0.921 | 0 | 9.38 | -0.043 | -0.045 |
| LIWC Tone Pos | 6.986 | 4.917 | 0 | 31.03 | 0.039 | 0.034 |
| LIWC Verb | 15.177 | 5.054 | 0 | 36 | -0.09 | 0.119 |
| LIWC Visual | 0.775 | 1.351 | 0 | 10.81 | 0.009 | 0.001 |
| LIWC Want | 0.321 | 0.829 | 0 | 8.99 | -0.001 | -0.007 |
| LIWC Wordcount | 60.942 | 17.212 | 28 | 97 | -0.087 | -0.006 |
| LIWC We | 1.479 | 3.23 | 0 | 22.58 | 0.04 | 0.029 |
| LIWC Wellness | 0.117 | 0.584 | 0 | 9.09 | -0.001 | -0.018 |
| LIWC Work | 4.9 | 5.389 | 0 | 40 | 0.039 | -0.006 |
| LIWC Words per sentence | 15.624 | 6.985 | 3.47 | 97 | 0.014 | -0.059 |
| LIWC You | 0.987 | 1.863 | 0 | 16.67 | 0.082 | 0.009 |
| Part Of Speech CC | 3.646 | 1.772 | 0 | 21 | -0.042 | 0.056 |
| Part Of Speech CD | 0.668 | 0.998 | 0 | 16 | -0.041 | -0.039 |
| Part Of Speech DT | 4.359 | 2.361 | 0 | 17 | -0.036 | 0.06 |
| Part Of Speech EX | 0.038 | 0.201 | 0 | 2 | 0.007 | 0.016 |
| Part Of Speech FW | 0.019 | 0.167 | 0 | 7 | -0.012 | -0.03 |
| Part Of Speech IN | 6.393 | 3.068 | 0 | 22 | -0.074 | -0.001 |
| Part Of Speech JJ | 6.444 | 3.134 | 0 | 23 | -0.021 | -0.059 |
| Part Of Speech LS | 0 | 0.012 | 0 | 1 | -0.007 | -0.012 |
| Part Of Speech MD | 0.523 | 0.833 | 0 | 7 | -0.011 | 0.021 |
| Part Of Speech NN | 18.628 | 6.965 | 3 | 51 | -0.024 | -0.076 |
| Part Of Speech PD | 0.048 | 0.229 | 0 | 2 | 0.001 | 0.031 |
| Part Of Speech PO | 0.092 | 0.339 | 0 | 6 | -0.007 | -0.002 |
| Part Of Speech PR | 3.171 | 2.373 | 0 | 20 | -0.014 | 0.022 |
| Part Of Speech RB | 3.026 | 2.459 | 0 | 20 | -0.124 | -0.01 |
| Part Of Speech RP | 0.254 | 0.538 | 0 | 4 | -0.066 | 0.006 |
| Part Of Speech SY | 0.003 | 0.053 | 0 | 2 | 0.01 | -0.005 |
| Part Of Speech TO | 1.992 | 1.508 | 0 | 13 | -0.013 | 0.056 |
| Part Of Speech UH | 0.011 | 0.11 | 0 | 3 | -0.019 | 0 |
| Part Of Speech VB | 11.998 | 4.558 | 0 | 29 | -0.135 | 0.068 |
| Part Of Speech WD | 0.194 | 0.474 | 0 | 6 | 0.018 | 0.038 |
| Part Of Speech WP | 0.286 | 0.599 | 0 | 5 | -0.018 | 0.016 |
| Part Of Speech WR | 0.218 | 0.503 | 0 | 4 | -0.025 | 0.018 |
| Contains List | 2.25 | 2.125 | 0 | 26 | 0.024 | -0.093 |
| Number Negations | 0.165 | 0.449 | 0 | 5 | -0.055 | -0.013 |
| Number Of Addresses | 0.003 | 0.053 | 0 | 1 | 0.005 | 0.021 |
| Number Of Names | 0 | 0.012 | 0 | 1 | 0.024 | 0.012 |
| Number Of Numbers | 0.783 | 1.559 | 0 | 30 | -0.006 | -0.034 |
| Number Of Punctuation | 8.255 | 5.18 | 0 | 174 | -0.019 | -0.043 |
| Number Of Question Marks | 0.04 | 0.277 | 0 | 9 | -0.026 | -0.005 |
| Number Of Symbols | 0.108 | 1.48 | 0 | 107 | 0.005 | -0.005 |
| URL Count | 0.004 | 0.083 | 0 | 4 | 0.001 | -0.021 |
| Flesch Kincaid Grade Level | 7.363 | 3.386 | 0 | 32.9 | 0.088 | -0.113 |
| Flesch Reading Ease Level | 69.988 | 16.841 | -23.45 | 111.78 | -0.119 | 0.129 |
| Sentiment AFINN | 8.642 | 6.309 | -16 | 44 | 0.014 | 0.047 |
| Sentiment NRC Anger | 0.009 | 0.019 | 0 | 0.22 | -0.018 | -0.032 |
| Sentiment NRC Anticipation | 0.077 | 0.054 | 0 | 0.316 | -0.005 | 0.01 |
| Sentiment NRC Disgust | 0.007 | 0.017 | 0 | 0.22 | 0.002 | -0.023 |
| Sentiment NRC Fear | 0.012 | 0.023 | 0 | 0.22 | 0.007 | -0.049 |
| Sentiment NRC Joy | 0.107 | 0.076 | 0 | 0.5 | -0.016 | 0.049 |
| Sentiment NRC Negative | 0.021 | 0.03 | 0 | 0.304 | -0.012 | -0.055 |
| Sentiment NRC Positive | 0.195 | 0.085 | 0 | 0.571 | 0.06 | 0.036 |
| Sentiment NRC Sadness | 0.017 | 0.026 | 0 | 0.222 | -0.016 | -0.014 |
| Sentiment NRC Surprise | 0.025 | 0.032 | 0 | 0.286 | 0.004 | -0.012 |
| Sentiment NRC Trust | 0.093 | 0.062 | 0 | 0.429 | 0.061 | -0.001 |
| Sentiment Polarity | 0.262 | 0.161 | -0.443 | 1 | 0.025 | 0.018 |
| Sentiment Subjectivity | 0.51 | 0.148 | 0 | 1 | -0.003 | 0.005 |
| Sentiment Vader | 0.812 | 0.265 | -0.895 | 0.998 | -0.021 | 0.015 |
| Lexical Diversity | 0.755 | 0.079 | 0.167 | 1 | -0.016 | -0.202 |
| Character Count | 341.203 | 107.151 | 126 | 705 | -0.025 | -0.058 |
| Contractions Count | 1.021 | 1.439 | 0 | 12 | -0.152 | 0.02 |
| Line Break Count | 0.986 | 1.76 | 0 | 26 | 0.05 | 0.041 |
| Longest Repetition Length | 1.973 | 1.249 | 1 | 45 | 0.071 | 0.126 |
| Mean Sentence Length | 16 | 7.546 | 3.737 | 89 | 0.01 | -0.066 |
| Mean Word Length | 4.565 | 0.559 | 3.265 | 7.933 | 0.142 | -0.157 |
| Number Of Exclamation Marks | 0.39 | 0.843 | 0 | 21 | -0.033 | -0.03 |
| Number Of Unique Words | 46.039 | 11.648 | 15 | 77 | -0.113 | -0.089 |
| Percentage Common 2-grams | 0.048 | 0.055 | 0 | 0.385 | -0.046 | 0.106 |
| Percentage Common 3-grams | 0.029 | 0.046 | 0 | 0.375 | -0.025 | 0.113 |
| Percentage Common 4-grams | 0.011 | 0.043 | 0 | 1 | -0.039 | 0.092 |
| Percentage Common Words | 0.156 | 0.096 | 0 | 0.688 | -0.04 | 0.12 |
| Percentage Rare 2-grams | 0.691 | 0.153 | 0 | 1 | 0.082 | -0.207 |
| Percentage Rare Words | 0.065 | 0.066 | 0 | 0.529 | 0.069 | -0.223 |
| Percentage Stop Words | 0.476 | 0.075 | 0 | 0.733 | -0.127 | 0.181 |
| Word Density | 0.183 | 0.018 | 0.112 | 0.241 | -0.159 | 0.151 |
| LDA Topic Vectors | Various techniques incl. structural topic models were explored but not used due to robustness and interpretability issues. | | | | | |

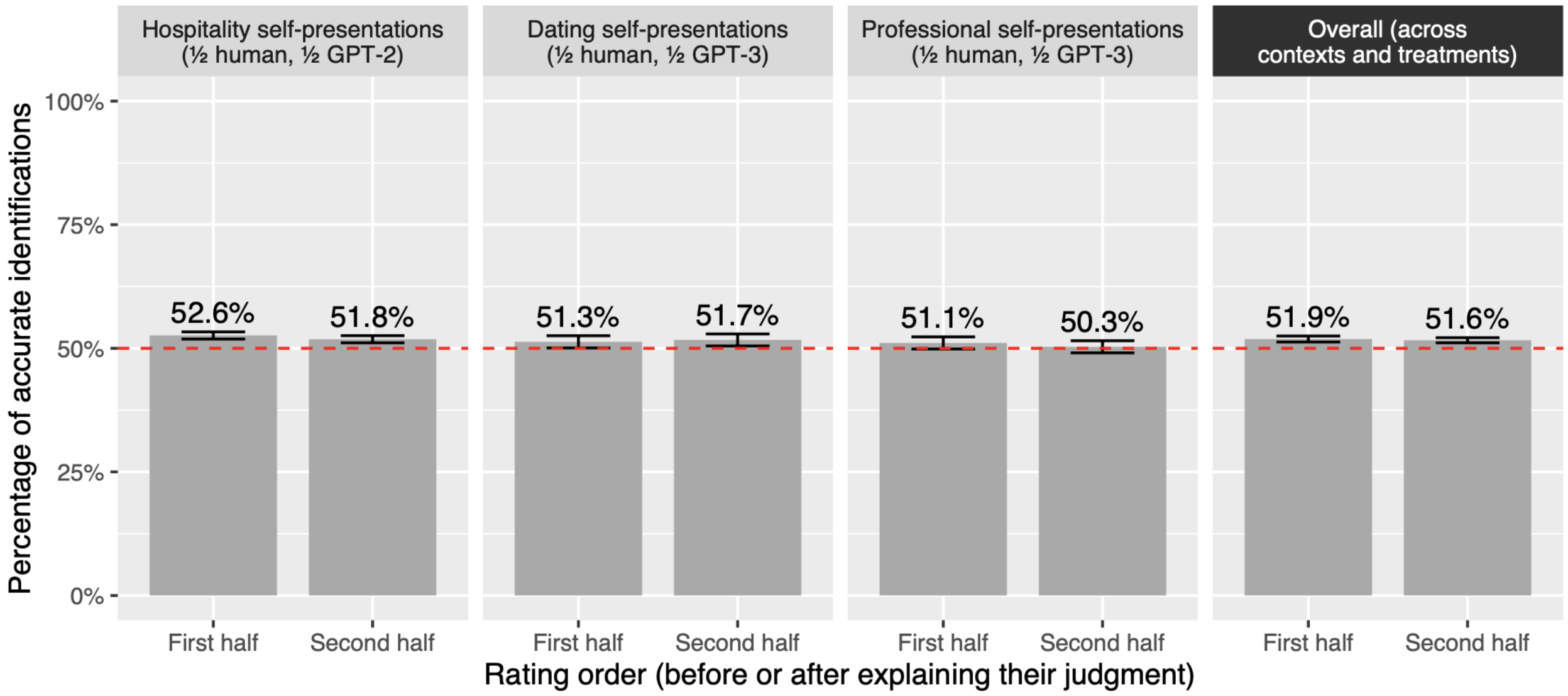

***Figure S1.*** *Participants' performance in identifying generated self-presentations did not change throughout the experiment. Error bars represent 95% confidence intervals for 6,000–16,000 judgments of 2,000–3,000 self-presentations per bar.*

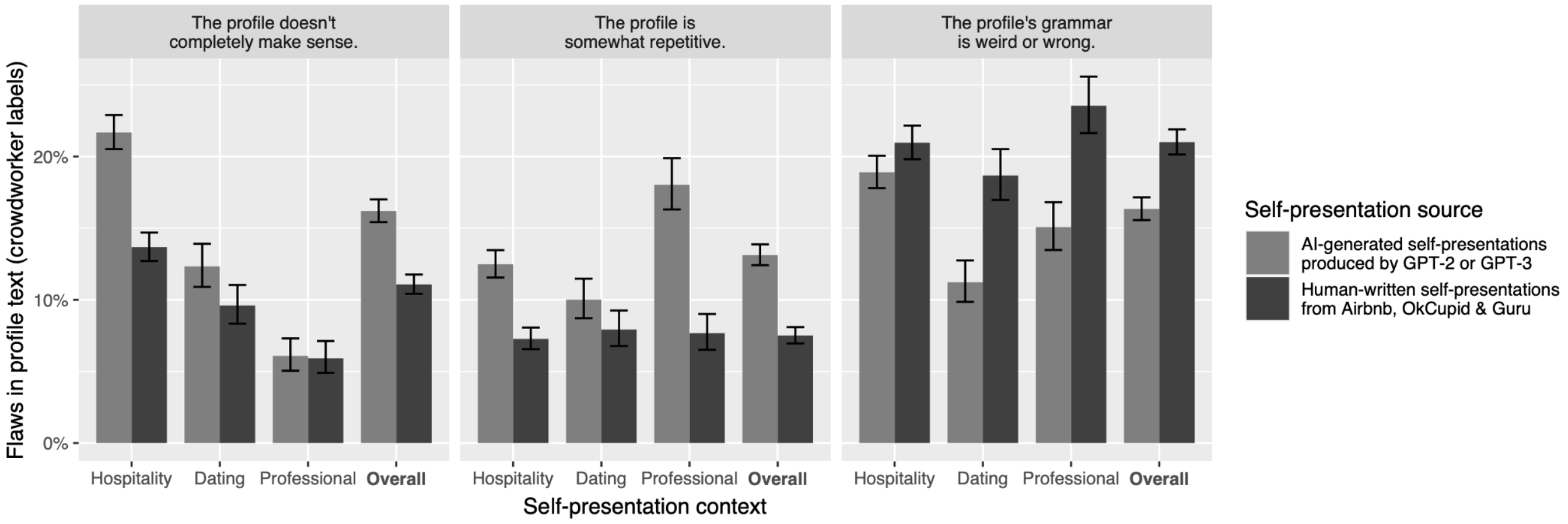

***Figure S2.*** Participants in a separate labeling task rated AI-generated self-presentations as nonsensical and repetitive more often than human-written self-presentations. *Error bars represent 95% confidence intervals for 1,898–4,704 judgments of 1,000–1,500 self-presentations per bar.*